\PassOptionsToPackage{unicode}{hyperref}
\PassOptionsToPackage{hyphens}{url}
\documentclass[11pt]{article}
\usepackage[margin=1in]{geometry}
\usepackage{xcolor}
\usepackage{amsmath,amssymb}
\usepackage{iftex}
\ifPDFTeX
  \usepackage[T1]{fontenc}
  \usepackage[utf8]{inputenc}
  \usepackage{textcomp} 
\else 
  \usepackage{unicode-math} 
  \defaultfontfeatures{Scale=MatchLowercase}
  \defaultfontfeatures[\rmfamily]{Ligatures=TeX,Scale=1}
\fi
\usepackage{lmodern}
\ifPDFTeX\else
\fi
\IfFileExists{upquote.sty}{\usepackage{upquote}}{}
\IfFileExists{microtype.sty}{
  \usepackage[]{microtype}
  \UseMicrotypeSet[protrusion]{basicmath} 
}{}
\makeatletter
\@ifundefined{KOMAClassName}{
  \IfFileExists{parskip.sty}{%
    \usepackage{parskip}
  }{
    \setlength{\parindent}{0pt}
    \setlength{\parskip}{6pt plus 2pt minus 1pt}}
}{
  \KOMAoptions{parskip=half}}
\makeatother
\usepackage{longtable,booktabs,array}
\usepackage{caption}
\AtBeginEnvironment{longtable}{\small}
\usepackage{calc} 
\usepackage{etoolbox}
\makeatletter
\patchcmd\longtable{\par}{\if@noskipsec\mbox{}\fi\par}{}{}
\makeatother
\IfFileExists{footnotehyper.sty}{\usepackage{footnotehyper}}{\usepackage{footnote}}
\makesavenoteenv{longtable}
\usepackage{graphicx}
\makeatletter
\newsavebox\pandoc@box
\newcommand*\pandocbounded[1]{
  \sbox\pandoc@box{#1}%
  \Gscale@div\@tempa{\textheight}{\dimexpr\ht\pandoc@box+\dp\pandoc@box\relax}%
  \Gscale@div\@tempb{\linewidth}{\wd\pandoc@box}%
  \ifdim\@tempb\p@<\@tempa\p@\let\@tempa\@tempb\fi
  \ifdim\@tempa\p@<\p@\scalebox{\@tempa}{\usebox\pandoc@box}%
  \else\usebox{\pandoc@box}%
  \fi%
}
\def\fps@figure{htbp}
\makeatother
\usepackage{bookmark}
\IfFileExists{xurl.sty}{\usepackage{xurl}}{} 
\hypersetup{
  pdftitle={KORA: Deterministic-First Execution Control for AI Workloads},
  hidelinks,
  pdfcreator={LaTeX via pandoc}}

\title{KORA: Deterministic-First Execution Control for AI Workloads}
\author{Albert Heekwan Kim\\Krako Labs\\\texttt{a.kim@krako.xyz}}
\date{}

\begin{document}
\maketitle

\section{Abstract}\label{abstract}

AI applications often invoke a model before they determine whether inference is necessary. That default is useful for ambiguous generative tasks, but it is poorly matched to work that is structured, deterministic, policy-bound, lookup-oriented, or directly verifiable. KORA introduces deterministic-first execution control: a pre-inference control layer that represents requests as task paths, attempts deterministic or structured routes first, validates route outcomes, records telemetry, and escalates to a model-candidate path only when deterministic handling is insufficient.

This paper presents KORA as a deterministic-first execution-control architecture and reports bounded evidence from reproducible benchmark and validation workloads. In a reproducible 100-task deterministic-heavy benchmark, KORA-controlled execution avoided 80 of 100 simulated model invocations versus a naive direct baseline, with zero mismatches in the deterministic subset. The workload is intentionally deterministic-heavy: it includes categories such as FAQ lookup, policy/rule checking, schema or format validation, deterministic transformation, and a smaller set of model-needed ambiguous or generative tasks. A separate synthetic customer-support local/no-network validation workload demonstrates route measurement and avoided model-call accounting in an application-shaped setting. These results support bounded model-invocation reduction and validation visibility in documented workloads. They do not establish production behavior, real-provider outcomes, energy outcomes, API-cost reduction, or broad superiority.

Keywords: Deterministic-first execution; execution control; model invocation routing; AI systems; reproducible benchmark; local/no-network validation.

\section{1. Introduction}\label{introduction}

Many AI systems route each request to a model as the first unit of execution. A request enters the application, a prompt is built, a model produces output, and downstream code validates or formats the result. This model-first pattern is easy to adopt, but it makes inference the default path even when the requested work can be handled by known logic.

The problem is not that models are unnecessary. Models remain valuable for ambiguous, context-heavy, generative, and judgment-oriented work. The problem is that model-first execution blurs the boundary between work that requires inference and work that can be resolved deterministically. Password reset instructions, policy lookups, structured formatting, schema validation, routing decisions, and budget checks often have known rules or expected outputs. Treating all of them as model tasks makes the system harder to measure and harder to control.

KORA places an execution-control layer before inference. It represents incoming requests as task paths, attempts deterministic or structured handling where appropriate, validates route outcomes, records route telemetry, and escalates only when deterministic handling is insufficient. The goal is not to replace models; the goal is to make model invocation an explicit decision rather than the default.

\begin{center}
\emph{\textbf{Structure first. Inference second.}}
\end{center}
\vspace{\baselineskip}

This paper makes four contributions. First, it defines KORA's deterministic-first execution-control model for routing AI work before inference. Second, it positions KORA relative to model serving, workflow orchestration, agent frameworks, retrieval-augmented generation, local runtime systems, and artifact-evaluation practices. Third, it reports the current reproducible deterministic-heavy benchmark result: 80 of 100 simulated model invocations avoided versus a naive direct baseline, with zero deterministic-output mismatches in the documented workload. Fourth, it summarizes a synthetic customer-support local/no-network validation workload that measures avoided model-call events, deterministic routes, and model escalations without external providers, API keys, private data, or network calls.

The remainder of the paper introduces the problem, places KORA in related work, defines the execution-control architecture, explains deterministic-first routing and validation, reports bounded evidence, and states limitations.

\section{2. Problem and Motivation}\label{problem-and-motivation}

Model invocation is often treated as the universal substrate for AI work. That design hides a control decision: some requests need generation or model judgment, while others need routing, lookup, deterministic transformation, policy enforcement, or validation. If every request is sent directly to a model, the system loses a clear account of which work genuinely required inference.

This matters for measurement. A model-first system can report model outputs, but it cannot easily separate deterministic-route success, model-candidate cases, fallback behavior, and avoidable model-call events. Without that separation, the system has weak evidence about where inference was necessary.

KORA addresses this by making the call/no-call boundary explicit before inference. A KORA-controlled workflow asks whether a request can be routed through deterministic handlers, structured lookups, or policy/validation logic before it reaches a model-candidate path. When the deterministic route is known and verifiable, the route can complete without a model. When no deterministic route exists, or when validation fails, the request conservatively escalates to model inference.

This framing is intentionally narrow. KORA does not claim that all AI work can be deterministic. It claims that many workloads contain deterministic or structured portions that should be routed and validated before inference when their route is known.

\section{3. Background and Related Work}\label{background-and-related-work}

KORA is related to several systems areas, but its control point is different. Many related systems optimize model execution, organize workflows, coordinate agents, retrieve context, or run models locally. KORA focuses on deciding whether a request should reach inference at all.

\subsection{3.1 Model Serving and Inference Systems}\label{model-serving-and-inference-systems}

LLM serving systems optimize model inference after a model call is selected. vLLM and PagedAttention address efficient memory management for large language model serving {[}R01{]}. ONNX Runtime provides an official inference runtime and optimization stack {[}R03{]}.

KORA does not claim to improve serving internals, scheduling, batching, memory management, kernel performance, or model runtime efficiency. Those systems matter after a model invocation is needed. KORA focuses on the preceding control question: whether a request should reach inference at all.

\subsection{3.2 Workflow and DAG Orchestration}\label{workflow-and-dag-orchestration}

Workflow systems such as Apache Airflow describe computation in DAG-oriented workflows {[}R07{]}. Snakemake provides a workflow-engine model for reproducible pipelines {[}R08{]}. These systems show the value of explicit execution structure, dependency ordering, and reproducible task paths.

KORA borrows the discipline of explicit execution paths, but its scope is narrower and AI-specific. It applies task-path structure to model-call boundaries, deterministic route validation, model escalation, and aggregate counters. KORA is not a general-purpose workflow scheduler and does not claim to replace workflow or DAG systems.

\subsection{3.3 Agent, Tool-Routing, and Programmatic AI Workflows}\label{agent-tool-routing-and-programmatic-ai-workflows}

Graph-based and agent-oriented frameworks can coordinate state, tools, and execution graphs. LangGraph documents graph-based agent execution primitives {[}R04{]}. DSPy describes declarative language-model pipelines and programmatic composition around model calls {[}R05{]}.

Agentic and tool-using language model systems provide further context. ReAct combines reasoning traces with actions to let language models interact with external environments and tools {[}R11{]}. Toolformer studies how language models can learn to use external tools through a model-centered training procedure {[}R12{]}. PAL uses language models to generate programs that are executed by an interpreter {[}R13{]}.

KORA is adjacent to these systems but emphasizes a different default. Agent and programmatic AI systems often organize model-centered workflows, generated programs, or tool-use behavior. KORA uses task structure, deterministic handlers, structured lookups, policy checks, and validation boundaries before deciding whether model inference is needed. It is complementary to agent and programmatic AI systems rather than a replacement for them.

\subsection{3.4 Retrieval-Augmented Generation}\label{retrieval-augmented-generation}

Retrieval-augmented generation grounds model outputs by retrieving external knowledge before generation {[}R06{]}. RAG is useful when a model still needs to generate or reason with retrieved context.

KORA is not a RAG replacement. RAG improves context selection for model use. KORA decides whether a task needs inference at all. A future system could use both: KORA could route deterministic work without inference and escalate retrieval-grounded questions to a RAG path when model reasoning is required.

\subsection{3.5 Local Runtime Systems}\label{local-runtime-systems}

Local runtime systems make it possible to run inference locally. The llama.cpp project provides local LLM inference in C/C++ {[}R09{]}. ONNX Runtime is also relevant as an inference runtime {[}R03{]}.

This paper uses local/no-network validation for current public evidence, but it does not present local model runtime validation as complete. Local runtime references are included as future-work context only. KORA does not currently claim measured local model performance, local runtime superiority, or integration evidence from local runtimes.

\subsection{3.6 Benchmarking and Artifact Evaluation}\label{benchmarking-and-artifact-evaluation}

Benchmark and artifact practices matter because KORA's current evidence depends on reproducible commands, synthetic workloads, aggregate counters, and explicit claim boundaries. ACM Artifact Review and Badging provides a public framework for artifact and reproducibility expectations {[}R10{]}.

KORA has not undergone formal artifact evaluation or external validation. This paper uses artifact-evaluation references only to motivate transparent reporting: reproducible commands, clear workload boundaries, local/no-network labels, and explicit non-claims.

\section{4. KORA Execution-Control Architecture}\label{kora-execution-control-architecture}

KORA is an execution-control layer between request intake and model invocation. Its purpose is not to make inference faster after a call is selected; it is to make the call/no-call boundary explicit before inference. A task graph represents possible work paths for an incoming request. A deterministic route resolves a request through known logic, rules, templates, or validation checks. A structured lookup maps a request to a known lookup or bounded retrieval path. Policy and validation logic confirm whether the selected route and output properties are acceptable. Model escalation marks the boundary where deterministic or structured handling is no longer sufficient. A model-candidate path is the counted or executable model-call path used after deterministic handling fails or is unavailable. Telemetry records route decisions, validation outcomes, model escalations, and avoided model-call events.

KORA task paths are not produced by unconstrained semantic planning. In the current design, task paths can be produced through explicitly declared application policies, handler registries, endpoint-level routing metadata, or bounded rule-based matching. Deterministic-first execution therefore requires developer or system-design effort: eligible task classes, deterministic handlers, routing policies, and validation expectations must be defined before the system can route work safely. This effort is most justified in domains with repeated, bounded, policy-like, schema-like, lookup-like, or transformation-oriented work. KORA should not be read as automatically solving arbitrary task decomposition. Accordingly, this paper evaluates KORA as an execution-control and measurement architecture, not as a complete production runtime scheduler.

\section{5. Deterministic-First Execution Model}\label{deterministic-first-execution-model}

KORA's execution model is organized around route classes and counters. The model-candidate path denotes a counted or executable model-call path used when deterministic or structured handling is not sufficient. In the current offline benchmark evidence, this remains a simulated model-invocation counter rather than a real provider call. The term marks the boundary where KORA no longer claims deterministic handling.

An avoided model-call event occurs when the baseline would count a model call for a request, but the KORA-controlled path handles the request through a deterministic or structured route without model escalation.

\begin{figure}[htbp]
\centering
\includegraphics[width=\linewidth]{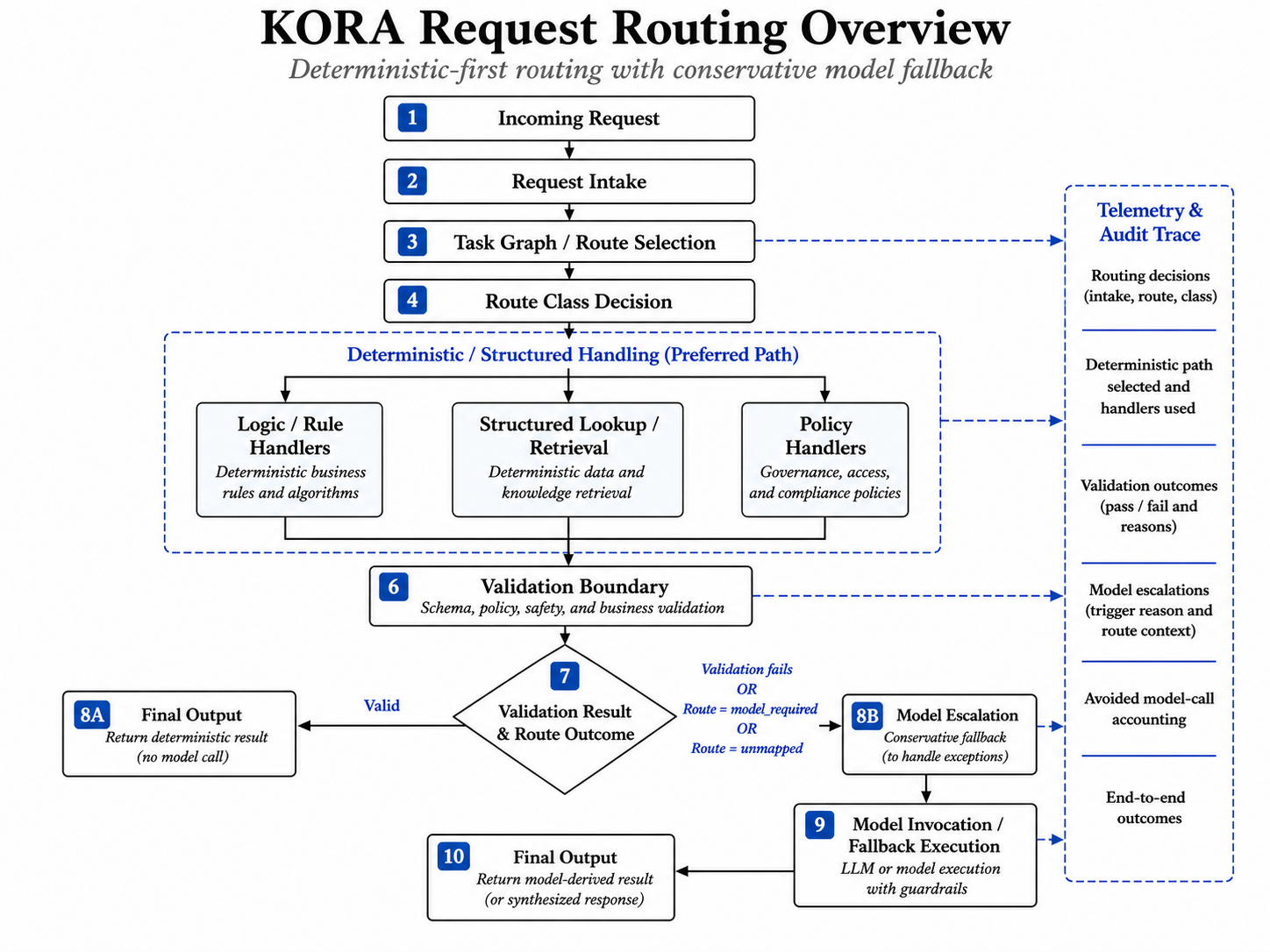}
\caption{KORA request-routing overview. Incoming requests are mapped to task paths and routed through deterministic, lookup, or policy handlers before model invocation. Validated deterministic outputs return without a model call, while failed, unmapped, or model-required routes conservatively escalate to model fallback. Telemetry records routing, validation, escalation, and avoided-call events.}
\label{fig:kora-request-routing-overview}
\end{figure}

\begin{figure}[htbp]
\centering
\includegraphics[width=\linewidth]{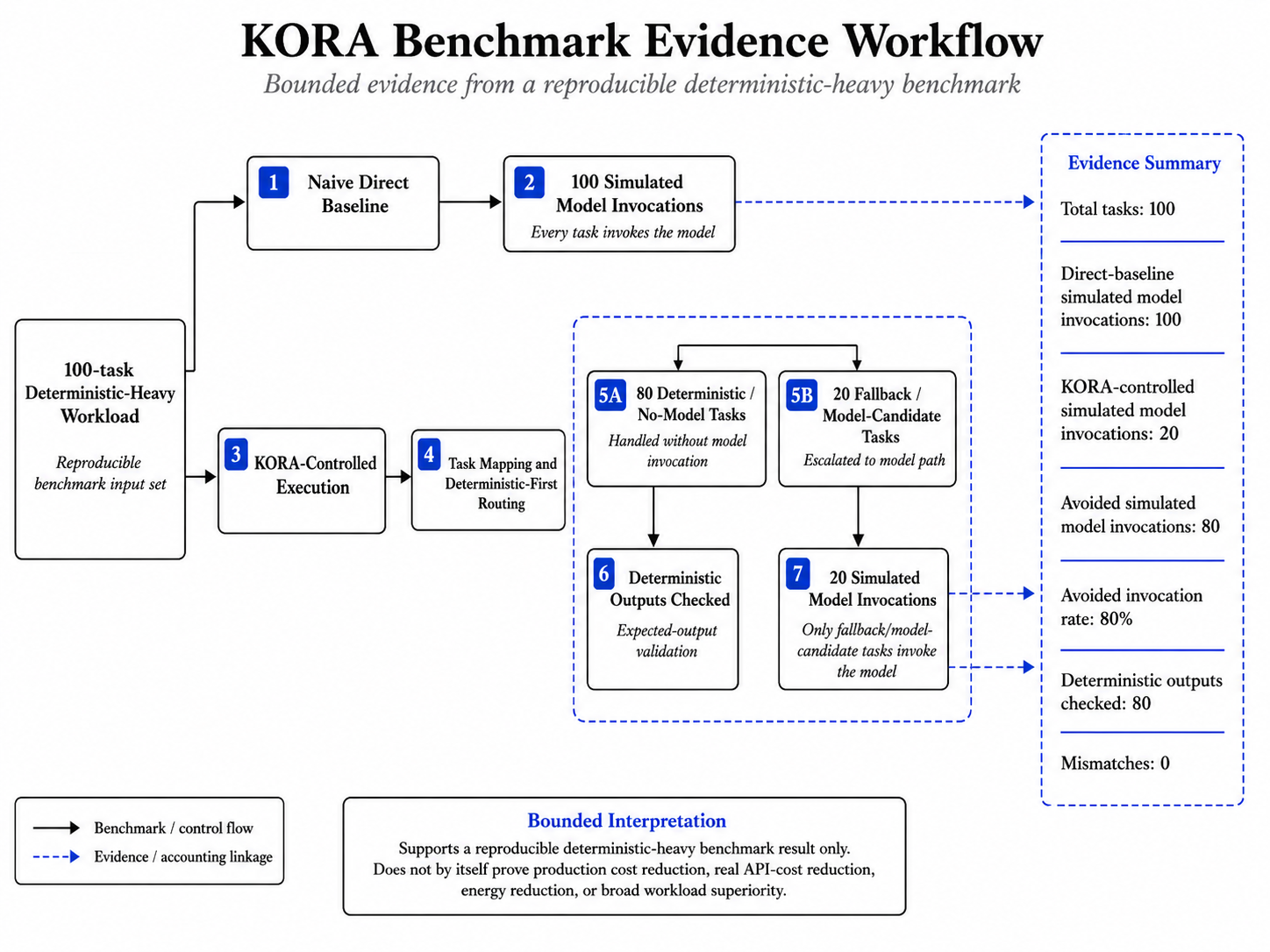}
\caption{KORA benchmark evidence workflow. The 100-task deterministic-heavy benchmark compares a naive direct baseline with KORA-controlled execution. KORA avoids 80 of 100 simulated model invocations while checking 80 deterministic outputs with zero mismatches; this result is bounded to the documented reproducible workload and is not production, cost, energy, or broad-superiority evidence.}
\label{fig:kora-benchmark-evidence-workflow}
\end{figure}

\begin{figure}[htbp]
\centering
\includegraphics[width=\linewidth]{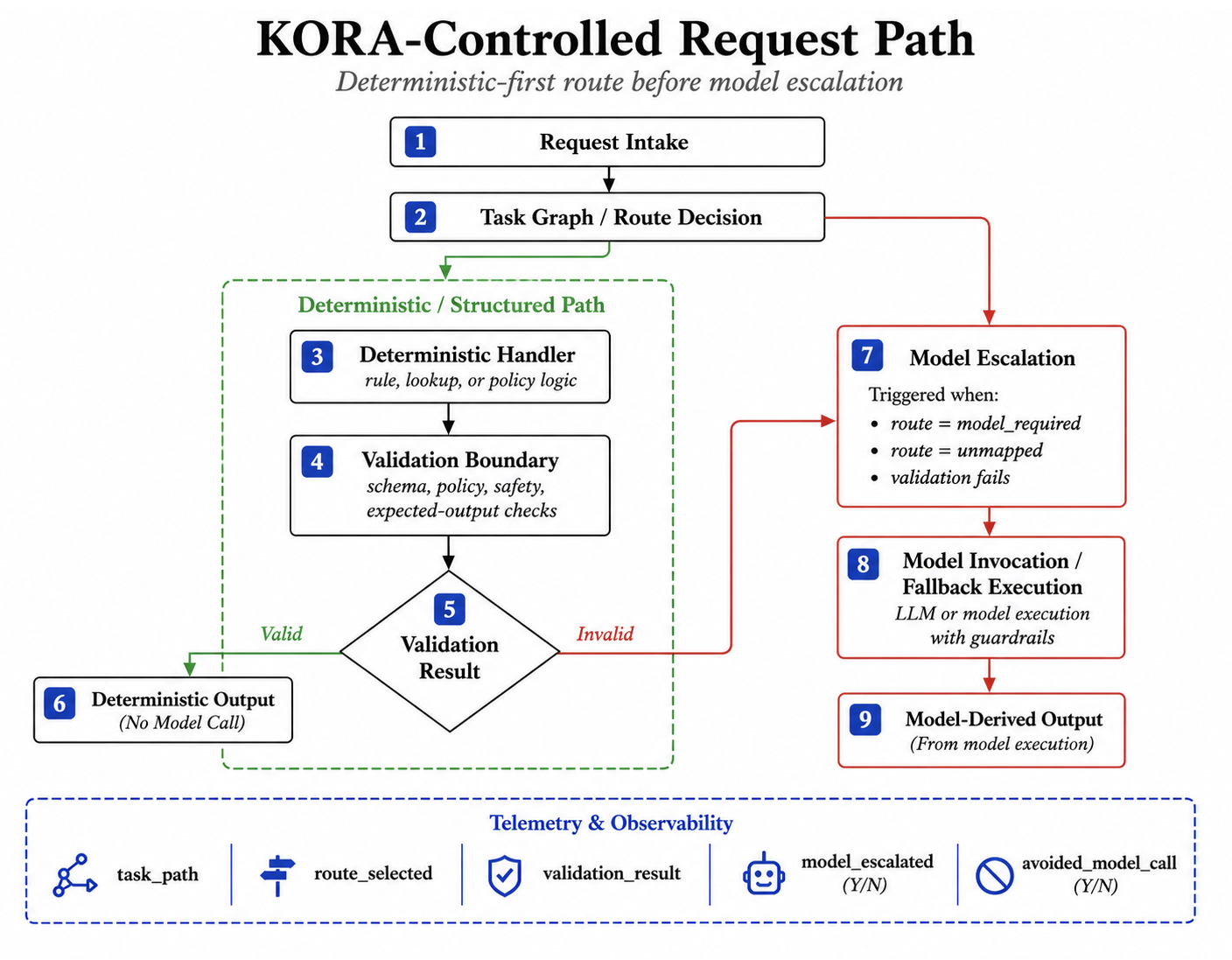}
\caption{KORA-controlled request path. KORA routes requests through deterministic or structured handling before model invocation, returning validated deterministic outputs when possible and conservatively escalating invalid, unmapped, or model-required routes to model fallback.}
\label{fig:kora-controlled-request-path}
\end{figure}

Current route classes include deterministic routes, structured lookups, policy routes, model-candidate paths, and model-required routes. A deterministic route uses known logic, templates, rules, or validations to handle the request. A structured lookup maps the request to a known structured handler or bounded lookup path. A policy route applies explicit policy logic or policy tables. A model-candidate path is used when deterministic handling is not sufficient, so the request is counted or prepared for model handling. A model-required route marks cases that should not be completed through deterministic handling alone.

The model-required label is conservative. It marks tasks that the configured policy, routing metadata, or handler registry does not consider eligible for deterministic resolution, as well as tasks explicitly classified as requiring open-ended generation, ambiguous reasoning, or model-backed interpretation. The label is not a claim of perfect classification. When the system cannot establish that a deterministic route is safe, conservative fallback is preferable to unsafe deterministic resolution.

The main metrics are baseline\_model\_calls, kora\_model\_calls, avoided\_model\_calls, avoided\_model\_call\_rate, deterministic\_routes, model\_escalations, and mismatch\_count. These metrics separate two questions: whether KORA reduced simulated model-call events, and whether validation outcomes remained expected for the workload under test.

{\def\LTcaptype{none} 
\begin{longtable}[]{@{}
  >{\centering\arraybackslash}p{(\linewidth - 2\tabcolsep) * \real{0.0745}}
  >{\raggedright\arraybackslash}p{(\linewidth - 2\tabcolsep) * \real{0.9255}}@{}}
\toprule\noalign{}
\begin{minipage}[b]{\linewidth}\centering
\textbf{Line}
\end{minipage} & \begin{minipage}[b]{\linewidth}\raggedright
\textbf{Pseudo-code}
\end{minipage} \\
\midrule\noalign{}
\endhead
\bottomrule\noalign{}
\endlastfoot
1 & Function KoraExecuteControl(R, H, V): \\
2 & // Counterfactual model-first baseline: each request would invoke the model. \\
3 & Telemetry.baseline\_model\_calls += 1 \\
4 & TaskPath \textless- SelectRoute(R, H) \\
5 & if TaskPath is Null or TaskPath.route\_class == "model\_required": \\
6 & return EscalateToModel(R) \\
7 & try: \\
8 & if TaskPath.route\_class == "deterministic\_route": \\
9 & Result \textless- ExecuteLogicHandler(R, TaskPath) \\
10 & else if TaskPath.route\_class == "structured\_lookup": \\
11 & Result \textless- ExecuteLookup(R, TaskPath) \\
12 & else if TaskPath.route\_class == "policy\_route": \\
13 & Result \textless- ExecutePolicy(R, TaskPath) \\
14 & else: \\
15 & return EscalateToModel(R) \\
16 & if ValidateOutput(Result, V{[}TaskPath{]}): \\
17 & Telemetry.deterministic\_routes += 1 \\
18 & Telemetry.avoided\_model\_calls += 1 \\
19 & return Result \\
20 & else: \\
21 & Telemetry.mismatch\_count += 1 \\
22 & return EscalateToModel(R) \\
23 & catch ExecutionError: \\
24 & return EscalateToModel(R) \\
25 & Function EscalateToModel(R): \\
26 & Telemetry.model\_escalations += 1 \\
27 & Telemetry.kora\_model\_calls += 1 \\
28 & O \textless- ExecuteModelCandidatePath(R) \\
29 & return O \\
\end{longtable}
}

\emph{\textbf{Algorithm 1.}} \emph{\textbf{KORA deterministic-first routing control.} The pseudo-code is conceptual and not implementation code. It shows task-path selection, deterministic or structured execution, validation before counting avoided model calls, and conservative escalation for unmapped, model-required, invalid, or failed routes.}

\section{\texorpdfstring{\hfill\break
6. Verification and Telemetry Framing}{ 6. Verification and Telemetry Framing}}\label{verification-and-telemetry-framing}

Verification is what makes deterministic-first routing reviewable. A deterministic route is useful only if expected output properties are known enough to check. KORA therefore treats validation as part of the route boundary: a deterministic handler should produce an output that can be checked against expected structure, expected values, or expected policy behavior.

Telemetry is what makes the route boundary auditable at workload scale. Counters record how many requests followed the direct baseline, how many followed KORA-controlled deterministic routes, how many escalated to model-candidate paths, and whether deterministic outputs matched expected results. These counters are evidence for the documented workloads; they are not production telemetry claims.

This distinction is central to the paper. KORA is not just an optimization layer. It is an execution-control layer that records when and why a request did or did not reach a model-candidate path.

\section{7. Evaluation Methodology}\label{evaluation-methodology}

This paper uses two current public workloads. The deterministic-heavy benchmark supports the primary quantitative claim. The customer-support triage workload demonstrates a synthetic local/no-network validation shape.

The deterministic-heavy benchmark contains 100 tasks. The workload is deliberately structured to stress the call/no-call boundary rather than to represent all possible AI workloads. Its task categories include FAQ-style lookup, policy or rule checking, schema and format validation, deterministic transformation, and a smaller model-needed class for ambiguous or generative tasks. The direct baseline counts one model call for each task. The KORA-controlled path routes deterministic work before inference and records model calls only for the remaining model-candidate routes.

This benchmark is appropriate first evidence because it isolates the central hypothesis: deterministic-heavy workloads should not require a model call for every request. The 80/20 split is therefore part of a controlled deterministic-heavy workload design. It should not be generalized to arbitrary workloads, production traffic, or broad AI-system behavior.

Figure 2 summarizes the benchmark evidence workflow by contrasting the direct baseline path with the KORA-controlled path under the documented 100-task deterministic-heavy workload.

The synthetic customer-support triage workload contains 12 synthetic requests across deterministic FAQ, deterministic policy, structured lookup, and model-required route classes. It is evaluated through local/no-network validation. This means it uses synthetic data, deterministic local validation paths, aggregate counters, and no external providers, API keys, private data, network calls, raw provider responses, or production traffic.

The workloads have different evidentiary roles. The deterministic-heavy benchmark supports the approved benchmark claim. The customer-support workload shows an application-shaped local/no-network validation path and demonstrates how KORA records avoided model-call events, deterministic routes, and model escalations. Neither workload establishes production behavior, real provider behavior, cost outcomes, energy outcomes, or broad superiority.

\section{8. Results}\label{results}

\subsection{8.1 Deterministic-heavy benchmark}\label{deterministic-heavy-benchmark}

Table 1 presents the confirmed benchmark composition and invocation-avoidance summary for the current reproducible deterministic-heavy workload.

{\def\LTcaptype{none} 
\begin{longtable}[]{@{}
  >{\centering\arraybackslash}p{(\linewidth - 8\tabcolsep) * \real{0.1464}}
  >{\centering\arraybackslash}p{(\linewidth - 8\tabcolsep) * \real{0.1609}}
  >{\centering\arraybackslash}p{(\linewidth - 8\tabcolsep) * \real{0.2215}}
  >{\raggedright\arraybackslash}p{(\linewidth - 8\tabcolsep) * \real{0.2424}}
  >{\raggedright\arraybackslash}p{(\linewidth - 8\tabcolsep) * \real{0.2289}}@{}}
\toprule\noalign{}
\begin{minipage}[b]{\linewidth}\centering
\textbf{Row group}
\end{minipage} & \begin{minipage}[b]{\linewidth}\centering
\textbf{Metric / category}
\end{minipage} & \begin{minipage}[b]{\linewidth}\centering
\textbf{Value}
\end{minipage} & \begin{minipage}[b]{\linewidth}\centering
\textbf{Interpretation}
\end{minipage} & \begin{minipage}[b]{\linewidth}\centering
\textbf{Boundary note}
\end{minipage} \\
\midrule\noalign{}
\endhead
\bottomrule\noalign{}
\endlastfoot
Workload composition & Total benchmark tasks & 100 & Controlled deterministic-heavy workload size & Not intended to represent all AI workloads \\
Workload composition & Deterministic / no-model tasks & 80 & Tasks eligible for deterministic or structured handling & Eligibility is defined by the benchmark workload \\
Workload composition & Fallback / model-candidate tasks & 20 & Tasks requiring model-candidate handling or fallback & Does not imply models are unnecessary \\
Workload categories & Category families included & FAQ/lookup-like; policy/rule checking; schema/format validation; deterministic transformation; model-needed ambiguous/generative tasks & Category families used to shape the workload & Category-level counts require workload-manifest verification \\
Baseline execution & Direct-baseline simulated model invocations & 100 & Naive model-first path counts one simulated invocation per task & Simulated benchmark counter, not real provider traffic \\
KORA-controlled execution & KORA-controlled simulated model invocations & 20 & Only model-candidate / fallback tasks count as simulated model invocations & Simulated benchmark counter \\
Invocation avoidance & Avoided simulated model invocations & 80 & Difference between baseline and KORA-controlled simulated invocation counts & Bounded to this deterministic-heavy workload \\
Invocation avoidance & Avoided invocation rate & 80\% & 80 avoided simulated invocations out of 100 baseline simulated invocations & Not a production cost-reduction claim \\
Validation / correctness & Deterministic outputs checked & 80 & Deterministic-path outputs were checked against expected outputs & Does not prove correctness for arbitrary production tasks \\
Validation / correctness & Mismatches & 0 & No mismatches in the deterministic subset under expected-output checks & Bounded to documented benchmark tasks \\
\end{longtable}
}

\emph{\textbf{Table 1. Deterministic-heavy benchmark composition and invocation-avoidance summary.} Category families are listed to describe the deterministic-heavy workload design; this table reports only the verified aggregate counts used for the current benchmark claim.}

The key quantitative result is as follows: in a reproducible 100-task deterministic-heavy benchmark workload, KORA-controlled execution avoided 80 of 100 simulated model invocations versus a naive direct baseline.

The interpretation is intentionally narrow. In this benchmark, KORA-controlled execution avoided 80 of 100 simulated model invocations versus a naive direct baseline while preserving expected deterministic outputs with zero mismatches. This is benchmark evidence for the documented deterministic-heavy workload. It is not production evidence, not real provider validation, and not a cost or energy claim.

\subsection{8.2 Customer-support triage synthetic workload}\label{customer-support-triage-synthetic-workload}

Table 2 summarizes the validation scope and evidence boundary of the synthetic customer-support triage workload. This workload supports route-measurement and avoided-call-accounting review; it does not provide production performance evidence.

{\def\LTcaptype{none} 
\begin{longtable}[]{@{}
  >{\centering\arraybackslash}p{(\linewidth - 8\tabcolsep) * \real{0.1459}}
  >{\centering\arraybackslash}p{(\linewidth - 8\tabcolsep) * \real{0.1343}}
  >{\centering\arraybackslash}p{(\linewidth - 8\tabcolsep) * \real{0.1892}}
  >{\raggedright\arraybackslash}p{(\linewidth - 8\tabcolsep) * \real{0.2726}}
  >{\raggedright\arraybackslash}p{(\linewidth - 8\tabcolsep) * \real{0.2580}}@{}}
\toprule\noalign{}
\begin{minipage}[b]{\linewidth}\centering
\textbf{Row group}
\end{minipage} & \begin{minipage}[b]{\linewidth}\centering
\textbf{Item}
\end{minipage} & \begin{minipage}[b]{\linewidth}\centering
\textbf{Current status}
\end{minipage} & \begin{minipage}[b]{\linewidth}\centering
\textbf{Interpretation}
\end{minipage} & \begin{minipage}[b]{\linewidth}\centering
\textbf{Boundary note}
\end{minipage} \\
\midrule\noalign{}
\endhead
\bottomrule\noalign{}
\endlastfoot
Workload scope & Workload type & Synthetic customer-support triage & Application-shaped validation workload & Not production customer-support traffic \\
Workload scope & Total requests & 12 & Small synthetic request set for route-coverage validation & Illustrative only, not broad benchmark evidence \\
Data boundary & Data source & Synthetic requests & No private user data or production tickets used & No real customer-support dataset claim \\
Execution boundary & Provider / network mode & Local / no-network & No external provider calls, API keys, or network calls required & Not real-provider validation \\
Route coverage & Route families exercised & Deterministic FAQ, policy, structured lookup, and model-required routes & Multiple route classes exercised in one application-shaped workload & Exact per-route counts require source verification \\
Telemetry coverage & Counters exercised & Baseline model calls, KORA-controlled model calls, avoided model calls, deterministic routes, and model escalations & Demonstrates route measurement and avoided-call accounting structure & Supports transparent route-level accounting within the local/no-network validation scope \\
Evidence role & Supported use & Route measurement and local validation & Shows how KORA records deterministic routes, avoided calls, and model escalations in a small synthetic workflow & Does not establish production benchmark performance \\
Non-claim boundary & Unsupported interpretation & Production customer-support performance, real-provider behavior, real API-cost reduction, or production cost reduction & These outcomes are outside the current local/no-network evidence & Must not be claimed from this workload \\
\end{longtable}
}

\emph{\textbf{Table 2. Customer-support triage local/no-network validation scope and boundary.} The table reports validation scope and evidence boundaries rather than final per-route performance claims.}

This is local/no-network validation only and not real provider validation. The workload uses synthetic customer-support requests to exercise deterministic FAQ, policy, structured lookup, and model-required route classes. The example shows that KORA can measure avoided model-call events in a synthetic application-shaped workflow, without API keys, external providers, private data, or network calls. The workload is useful for route-coverage review, but it is not a production customer-support benchmark.

\subsection{8.3 Latency direction}\label{latency-direction}

This paper does not include measured latency evidence yet. The expected direction is route-dependent: deterministic fast paths can reduce latency by avoiding inference; structured lookup and policy routes can avoid model wait time when validation succeeds; and model escalation paths may add orchestration overhead before inference.

From an engineering perspective, deterministic routing, handler lookup, and expected-output verification are expected to be lightweight relative to model inference for many structured tasks, especially when the route is local and the verification target is explicit. That expectation is a design rationale, not a measured production result. The actual overhead depends on handler complexity, validation depth, orchestration design, and whether a request ultimately escalates to a model-candidate path.

One simple explanatory model is:

\[
T_{\mathrm{total}} =
T_{\mathrm{routing}} +
T_{\mathrm{validation}} +
(1 - P_{\mathrm{avoid}}) \cdot T_{\mathrm{inference}}
\]

Here, \emph{T\textsubscript{routing}} denotes routing or policy-decision overhead, \emph{T\textsubscript{validation}} denotes deterministic verification or expected-output checking overhead, \emph{P\textsubscript{avoid}} denotes the fraction of tasks resolved without model invocation in a given workload, and \emph{T\textsubscript{inference}} denotes a model-invocation latency or cost proxy. This model is explanatory only. If fallback occurs, KORA can add routing and validation overhead before model invocation. The design assumption is that routing and validation overhead are lightweight relative to model inference for eligible deterministic tasks, but this remains an engineering expectation until measured in runtime-specific latency benchmarks.

Measured p50/p95 latency remains future work. Until those measurements exist, KORA should not be described as universally faster across all workloads.

\section{9. Limitations and Non-Claims}\label{limitations-and-non-claims}

Current limitations include synthetic workloads, limited workload diversity, no production traffic, no real provider validation, no measured provider-cost outcome, no production benchmark evidence, no full runtime-integrated real-provider benchmark evidence, no measured latency distribution, no energy measurement, no user study, and no broad superiority claim.

KORA requires upfront engineering effort: deterministic handlers, routing policies, and validation rules must be designed before repeated bounded work can bypass inference. In domains with recurring policy-like, schema-like, lookup-like, or format-constrained tasks, that effort may be more predictable and auditable than ongoing prompt tuning or hallucination debugging in model-first systems.

Table 3 separates supported findings from unsupported interpretations.

{\def\LTcaptype{none} 
\begin{longtable}[]{@{}
  >{\raggedright\arraybackslash}p{(\linewidth - 4\tabcolsep) * \real{0.2757}}
  >{\raggedright\arraybackslash}p{(\linewidth - 4\tabcolsep) * \real{0.3905}}
  >{\raggedright\arraybackslash}p{(\linewidth - 4\tabcolsep) * \real{0.3338}}@{}}
\toprule\noalign{}
\begin{minipage}[b]{\linewidth}\centering
\textbf{Evidence item}
\end{minipage} & \begin{minipage}[b]{\linewidth}\centering
\textbf{Supported interpretation}
\end{minipage} & \begin{minipage}[b]{\linewidth}\centering
\textbf{Explicit non-claim / boundary}
\end{minipage} \\
\midrule\noalign{}
\endhead
\bottomrule\noalign{}
\endlastfoot
Reproducible 100-task deterministic-heavy benchmark & Supports bounded evaluation of deterministic-first execution on a controlled deterministic-heavy workload & Does not support broad workload superiority across arbitrary AI workloads \\
100 direct-baseline simulated model invocations & Shows the naive model-first comparison path used for the benchmark & Does not imply real provider traffic or production request volume \\
20 KORA-controlled simulated model invocations & Shows that only model-candidate / fallback tasks counted as simulated model invocations under KORA-controlled execution & Does not imply that KORA eliminates the need for models \\
80 avoided simulated model invocations & Supports invocation-avoidance evidence under the documented deterministic-heavy benchmark conditions & Does not establish production cost reduction or real API-cost reduction \\
80\% avoided invocation rate & Summarizes the avoided simulated invocation rate for the benchmark workload & Does not imply the same rate for production, general workloads, or unseen task distributions \\
80 deterministic outputs checked & Shows that deterministic-path outputs were checked against expected outputs in the benchmark & Does not prove correctness for arbitrary production tasks \\
0 mismatches in the deterministic subset & Supports expected-output consistency for the documented deterministic subset & Does not prove universal correctness, safety, or reliability \\
20 fallback / model-candidate tasks & Shows conservative fallback behavior for tasks not handled deterministically & Does not claim that all tasks can be deterministically resolved \\
Telemetry / counter traces & Supports reproducible interpretation of routing, validation, fallback, and avoided-call accounting & Does not prove production telemetry deployment or runtime observability at scale \\
Synthetic customer-support local/no-network validation & Illustrates route measurement and local validation in an application-shaped setting & Does not establish production customer-support performance or real-provider behavior \\
Local/no-network validation setup & Shows that the reported validation path does not require external providers, API keys, private data, or network calls & Does not establish real provider outcomes, API billing outcomes, or production runtime behavior \\
KORA execution-control framing & Supports the claim that KORA makes model invocation an explicit routing decision before inference & Does not claim superiority over model-serving systems, workflow orchestrators, agent frameworks, RAG systems, or local runtimes \\
\end{longtable}
}

\emph{\textbf{Table 3. Evidence boundary and explicit non-claims.} The table records supported interpretations and corresponding non-claims side by side.}

The deterministic-heavy benchmark and customer-support triage workload are useful early evidence, but they are not sufficient for production, cost, energy, or broad generalization claims. They show that KORA can avoid or measure model-call events in documented deterministic-heavy and synthetic local/no-network validation contexts.

The related-work references integrated in this paper provide background and positioning. They do not prove KORA's benchmark claims or establish superiority over cited systems.

\section{10. Discussion}\label{discussion}

KORA's main contribution is the placement of execution control before model invocation. This changes how an AI system can be evaluated. Instead of asking only how well a model responds, the system can ask whether a model call was needed, whether a deterministic route was valid, and whether the fallback boundary was exercised.

This framing is useful for deterministic-heavy workloads because it makes model invocation measurable rather than automatic. It also helps separate route correctness from model quality. If a deterministic handler is expected to produce a known output, the benchmark can check that output directly. If deterministic handling is not appropriate, the request can be counted as a model-candidate path without pretending that deterministic execution solved it.

The design does not eliminate the need for models. It makes model use more explicit. In future work, KORA can be combined with model-serving systems, workflow orchestration, RAG, local inference runtimes, or agent frameworks. The current paper does not claim that KORA is superior to those systems; it identifies a complementary control layer around the call boundary.

\section{11. Reproducibility and Artifact Notes}\label{reproducibility-and-artifact-notes}

KORA's current public evidence is organized around reproducible commands, local/no-network examples, aggregate counters, and claim-safe reports. This aligns with systems-paper artifact discipline but does not constitute formal artifact evaluation. ACM artifact review and badging guidance is cited only as a reference point for transparent artifact practices {[}R10{]}.

Reproduction requires four checks: run the deterministic-heavy benchmark in offline mode; run the customer-support triage local/no-network validation example; inspect aggregate counters and claim-boundary text; and keep raw provider responses, credentials, private data, and production traffic out of public artifacts.

\section{12. Conclusion}\label{conclusion}

Model-first execution is not always necessary. Many AI workloads include structured, deterministic, policy-driven, or validation-oriented tasks that can be routed before inference.

KORA introduces deterministic-first execution control: it represents requests as structured task paths, attempts deterministic handling before inference, validates route outcomes, records telemetry, and escalates to a model-candidate path only when needed. Current evidence shows avoided simulated model invocations in a reproducible deterministic-heavy benchmark and measurable avoided model-call events in a synthetic customer-support local/no-network validation workload.

KORA is complementary to model-serving systems, workflow orchestrators, agent frameworks, RAG systems, and local runtimes. Future work should expand runtime validation, latency measurement, local model evaluation, and real-provider experiments without weakening claim boundaries.

The repository contains the implementation and public documentation used to support the reproducible benchmark workflow described in this paper. The current evidence does not claim formal artifact evaluation, production deployment validation, real-provider cost reduction, or completeness of submission materials.

\section{Code and Artifact Availability}\label{code-and-artifact-availability}

KORA is available as an open-source project at: \url{https://github.com/Krako-Labs/KORA}

The repository contains the implementation, benchmark workflow, and public documentation associated with the reproducible evaluation described in this paper. The current results should be interpreted within the documented benchmark boundaries.

\section{References}\label{references}

{[}R01{]} Woosuk Kwon, Zhuohan Li, Siyuan Zhuang, Ying Sheng, Lianmin Zheng, Cody Hao Yu, Joseph E. Gonzalez, Hao Zhang, and Ion Stoica. ``Efficient Memory Management for Large Language Model Serving with PagedAttention.'' SOSP 2023; arXiv:2309.06180. Source: \url{https://arxiv.org/abs/2309.06180}; \url{https://doi.org/10.48550/arXiv.2309.06180}

{[}R03{]} ONNX Runtime project. ``ONNX Runtime'' official documentation. Source owner: ONNX Runtime project / Microsoft. n.d.; Accessed: May 2026; Source: \url{https://onnxruntime.ai/docs/}

{[}R04{]} LangChain. ``LangGraph Graph API overview.'' Official LangGraph documentation. n.d.; Accessed: May 2026; Source: \url{https://docs.langchain.com/oss/python/langgraph/graph-api}

{[}R05{]} Omar Khattab, Arnav Singhvi, Paridhi Maheshwari, Zhiyuan Zhang, Keshav Santhanam, Sri Vardhamanan, Saiful Haq, Ashutosh Sharma, Thomas T. Joshi, Hanna Moazam, Heather Miller, Matei Zaharia, and Christopher Potts. ``DSPy: Compiling Declarative Language Model Calls into Self-Improving Pipelines.'' arXiv:2310.03714, 2023. Source: \url{https://arxiv.org/abs/2310.03714}

{[}R06{]} Patrick Lewis, Ethan Perez, Aleksandra Piktus, Fabio Petroni, Vladimir Karpukhin, Naman Goyal, Heinrich Kuttler, Mike Lewis, Wen-tau Yih, Tim Rocktaschel, Sebastian Riedel, and Douwe Kiela. ``Retrieval-Augmented Generation for Knowledge-Intensive NLP Tasks.'' NeurIPS 2020; arXiv:2005.11401. Source: \url{https://arxiv.org/abs/2005.11401}; \url{https://doi.org/10.48550/arXiv.2005.11401}

{[}R07{]} Apache Airflow project. ``Dags.'' Apache Airflow 3.2.1 documentation. Source owner: Apache Software Foundation. n.d.; Accessed: May 2026; Source: \url{https://airflow.apache.org/docs/apache-airflow/stable/core-concepts/dags.html}

{[}R08{]} Johannes Koster and Sven Rahmann. ``Snakemake-a scalable bioinformatics workflow engine.'' Bioinformatics 28(19), 2520-2522, 2012. Source: \url{https://doi.org/10.1093/bioinformatics/bts480}

{[}R09{]} ggml-org. ``llama.cpp: LLM inference in C/C++.'' Official GitHub repository. n.d.; Accessed: May 2026; Source: \url{https://github.com/ggml-org/llama.cpp}

{[}R10{]} Association for Computing Machinery. ``Artifact Review and Badging - Current.'' Version 1.1, 2020. Source: \url{https://www.acm.org/publications/policies/artifact-review-and-badging-current}

{[}R11{]} Shunyu Yao, Jeffrey Zhao, Dian Yu, Nan Du, Izhak Shafran, Karthik Narasimhan, and Yuan Cao. ``ReAct: Synergizing Reasoning and Acting in Language Models.'' ICLR 2023; arXiv:2210.03629. Source: \url{https://arxiv.org/abs/2210.03629}; \url{https://doi.org/10.48550/arXiv.2210.03629}

{[}R12{]} Timo Schick, Jane Dwivedi-Yu, Roberto Dessi, Roberta Raileanu, Maria Lomeli, Luke Zettlemoyer, Nicola Cancedda, and Thomas Scialom. ``Toolformer: Language Models Can Teach Themselves to Use Tools.'' arXiv:2302.04761, 2023. Source: \url{https://arxiv.org/abs/2302.04761}; \url{https://doi.org/10.48550/arXiv.2302.04761}

{[}R13{]} Luyu Gao, Aman Madaan, Shuyan Zhou, Uri Alon, Pengfei Liu, Yiming Yang, Jamie Callan, and Graham Neubig. ``PAL: Program-aided Language Models.'' arXiv:2211.10435, 2022; revised 2023. Source: \url{https://arxiv.org/abs/2211.10435}; \url{https://doi.org/10.48550/arXiv.2211.10435}

\end{document}